\documentclass[11pt,a4paper]{article}
\usepackage[hyperref]{emnlp2020}
\usepackage{times}
\usepackage{latexsym}

\usepackage{enumitem}
\usepackage{amsmath,empheq}
\usepackage{courier}
\usepackage{amsmath}

\usepackage{times}  
\usepackage{helvet}  
\usepackage{courier}  
\usepackage{url}  
\usepackage{graphicx}  
\usepackage{multirow}
\usepackage{latexsym}
\usepackage{wasysym}
\usepackage{bm}

\usepackage{microtype}

\aclfinalcopy 

\title{CRAB: Class Representation Attentive BERT for Hate Speech Identification in Social Media}

\author{Sayyed M. Zahiri \\
  Georgia Institute of Technology \\
  \texttt{mzahiri@gatech.edu} \\\And
  Ali Ahmadvand \\
  Emory University \\
  \texttt{aahmadv@emory.edu} \\}

\date{}

\begin{document}
\maketitle
\begin{abstract}

In recent years, social media platforms have hosted an explosion of hate speech and objectionable content. The urgent need for effective automatic hate speech detection models have drawn remarkable investment from companies and researchers. Social media posts are generally short and their semantics could drastically be altered by even a single token. Thus, it is crucial for this task to learn context-aware input representations, and consider relevancy scores between input embeddings and class representations as an additional signal. To accommodate these needs, this paper introduces CRAB (Class Representation Attentive BERT), a neural model for detecting hate speech in social media. The model benefits from two semantic representations: (i) trainable token-wise and sentence-wise class representations, and  (ii) contextualized input embeddings from state-of-the-art BERT encoder. To investigate effectiveness of CRAB, we train our model on Twitter data and compare it against strong  baselines. Our results show that CRAB achieves 1.89\% relative improved Macro-averaged F1 over state-of-the-art baseline. The results of this research open an opportunity for the future research on automated abusive behavior detection in social media.



\end{abstract}

\section{Introduction and Related Work}


\noindent Twitter is one of the popular social media platforms in which people post several hundred million tweets on daily basis. Twitter similar to other existing social networks, greatly suffers from the range of violence, hate speech and human right abuse imposed on specific groups or individuals~\cite{founta2018large}. Hence, it is imperative to protect the users by taking pro-active steps and develop algorithms to automatically identify hate messages, and prevent them from spreading. 

Essentially, there are two steps associated with automatic hate speech detection task: (i) Annotated data collection (ii) Model development. For the first step, leveraging crowd-sourcing is one of the most common approaches; for the second one, researchers have leveraged variety of Natural Language Processing (NLP) techniques. \citeauthor{pereira2019detecting} gathered annotated tweets through crowd-sourcing and introduced a social network analyzer which allows researchers monitor hate speech in tweets. The authors formulated abusive tweet identification as a text classification problem and developed several NLP techniques to accomplish this goal. Similarly, in this paper, we tackle hate speech detection task in the setting of text classification task.





Text classification is one of the fundamental NLP tasks used in social media analysis. Traditional text classification task mainly relies on vector space models created on hand-crafted features such as Term Frequency-Inverse Document Frequency (TF-IDF) and n-gram \cite{zhang2011comparative,wang2012baselines}. \citeauthor{gaydhani2018detecting} applied TF-IDF feature extraction technique followed by traditional machine learning models on tweets to detect hate speech. Although these techniques have been effective in social media mining, they suffer from vocabulary mismatch and ambiguity~\cite{croft2010search}. Later, deep neural models such as Convolutional Neural Networks (CNN) and Recurrent Neural Networks (RNN) \cite{kim2014convolutional,zhang2015character,zahiri2017emotion,ravuri2015recurrent} have mitigated above-mentioned shortcomings by learning dense text representation with minimal hyper-parameter tuning. In the context of social media analysis,~\citeauthor{gamback2017using} utilized different variations of CNN based models to assign each tweet with one of the predefined labels.  One crucial limitation in these classifiers is they do not take class representations into consideration. \citeauthor{du2019explicit} resolved this limitation by introducing interaction mechanism which computes matching score between encoded input \cite{qiao2018new} and classes, then calculated scores were utilized to predict the class. 

\par More recently,~\citeauthor{vaswani2017attention} developed the transformer model by using stacked self-attention and fully connected layers. The authors demonstrated the effectiveness of this model in capturing long term dependencies from text sequences similar to recurrent networks. However,  owing to it's feed-forward architecture it could be trained in a more efficient way compared to typical RNNs. Inspired by this work,~\citeauthor{devlin2018bert} proposed BERT language model. BERT is a multi-layer bi-directional transformer trained on a very large-scaled unlabeled corpora to learn text representations. Fine-tuned BERT encoder has improved text classification task performance by a substantial margin on the benchmark dataset~\cite{sun2019fine}.

\begin{figure*}[htbp!]
\centering\small
\includegraphics[width = 15cm]{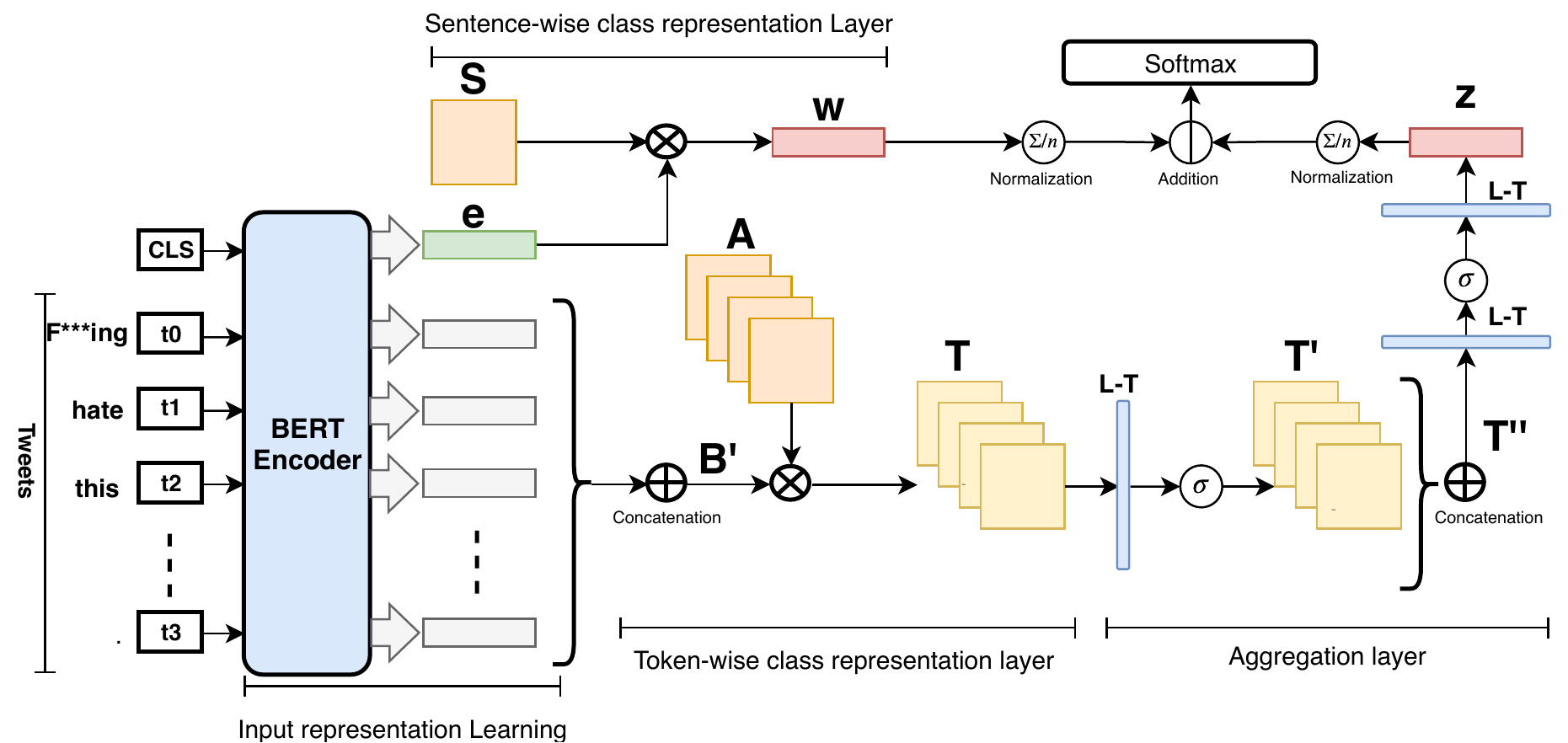}
\caption{Class Representation Attentive BERT (CRAB) overall architecture, L-T: Linear transformation.}
\label{fig:model}
\vspace{-2mm}
\end{figure*}



Although, state-of-the-art transformer-based models have shown promising results in text classification, similar to CNN and RNN based models, they do not incorporate information embedded in the class representations. Inspired by ~\citeauthor{du2019explicit}, we introduce CRAB, an
interaction-based classifier which relies on the similarity scores between encoded input and class representations. Our framework embraces three parts: \textit{input representation layer}, \textit{class representation layers}, and \textit{aggregation layer}. The input representation layer projects tweets into token-level and sentence-level dense embedding spaces. Class representation layers map classes
into latent representations and let the network interact with the encoded input and determine the similarity scores between them. In other words, these layers are trained to learn the matching scores between classes and each part of the input in an end to end fashion. Finally, aggregation layer combines the matching scores computed in previous layers and infer the class label. In this paper, we use the annotated Twitter data gathered by \cite{founta2018large}. In summary, the contributions of this paper are as follows:



\vspace{-0.2cm}

\begin{itemize}[noitemsep]
\item A new model which leverages matching scores between trainable class representations and encoded input data to detect hate speech.
\item We perform extensive experiments on Twitter data to show our proposed model outperforms several strong baselines.    
\end{itemize}

\section{Model Overview}

In this section we introduce our proposed model, CRAB. The overall architecture of CRAB is illustrated in Figure~\ref{fig:model}.
The objective of this model is to take the tweets and classify them into one of predefined classes (Multi-class Classification). More specifically, given the training set $D = \{(X_n,Y_n)\}^{N-1}_0$ ($X_n$ is the \textit{n-th} training example and $Y_n$ is its corresponding label structured as a one hot vector), the goal of the classifier is to learn $f : X \to Y$ such that empirical risk is minimized by \textit{N} observations: 

\begin{equation} 
\min_{f\in F}\frac{1}{N}\sum_{i=1}^{N}L(Y_i,f(X_i))
\end{equation}

Loss function \textit{L} is a continuous function that penalizes training error. In this work, we employ cross entropy loss function. CRAB is comprised of: (i) representation layer (section \ref{representation_layer}) (ii) token-wise class representation layer (section \ref{token-wise}) (iii) sentence-wise class representation layer (section \ref{Sentence-wise}) (iv) aggregation layer (section \ref{agg_lay}).


\subsection{Input Representation Layer}
\label{representation_layer}
The purpose of representation learning layer is to generate a contextualized fixed-sized embedding vectors for the tweets. We utilize BERT encoder to vectorize the input data. BERT can provide a more sophisticated text representation by learning from left and right of a token's context in all layers. CRAB takes all the BERT embedding generated from final block of the transformer. More precisely, this layer encodes the input tweets into matrix $B=(e_0,...,e_{N-1}),B\in R^{|k|\times N}$ where $e_{i\in\{0,...,N-1\}}$ is the embedding vector for the \textit{i-th} token, \textit{N} is input text length, and \textit{$|k|$} is embedding dimension.







\subsection{Token-wise Class Representation Layer}
\label{token-wise}
This layer is devised to allow the neural model to learn how the encoded classes should attend to every single token in the input. To this end, we introduce a multi-head token-wise class representation network \textbf{\textit{A}}. Each head in this block learns the interactions between the encoded input tokens and 
classes independently. This layer takes $B'=(e_1,...,e_{N-1})$ as an input; $B'$ 
is concatenation of all final layer's hidden states except the hidden state 
corresponds to the first token (special token [CLS]). Similar to previous studies \cite{du2019explicit}, this layer calculates the matching scores between classes and input data using dot product operation:
\begin{table}[h]
\centering\resizebox{\columnwidth}{!}{
\begin{tabular}{l||r|r|r|r||r}
\multicolumn{1}{c||}{\bf Class} & \multicolumn{1}{c|}{\bf Normal} & \multicolumn{1}{c|}{\bf Abusive} & \multicolumn{1}{c|}{\bf Spam} & \multicolumn{1}{c||}{\bf Hateful} & \multicolumn{1}{c}{\bf Total} \\
\hline\hline
Count & 53851 & 27150 & 14030 & 4965 & 99996 \\

\end{tabular}}
\caption{The total number of tweets per class.}
\label{tbl:class_dist}
\vspace{-2mm}
\end{table}

\begin{equation} 
T_i = A_i \times B', i \in\{0,...,m\}
\end{equation}

Where $A = (A_0,...,A_m)$, $A_i\in R^{c\times |k|}$ and $T= (T_0,...,T_m)$,$ T_i\in R^{c \times (N-1) }$;  number of classes is shown as \textit{c}, and \textit{m} is number of class representation heads.


\subsection{Sentence-wise Class Representation Layer} 
\label{Sentence-wise}
Sentence-wise trainable class representation matrix is depicted as $\textbf{S} \in R^{c\times |k|}$ in Figure~\ref{fig:model}. Given the sentence embedding, $\textbf{\textit{e}} \in R^{|k| \times 1}$ as an input to this layer, \textbf{S} is tuned to learn sentence-level class representation during the training process. Similar to sub-section~\ref{token-wise}, here, we also  apply dot product to compute matching scores (shown as \textbf{\textit{w}}) between the sentence embedding \textbf{\textit{e}}, and sentence-wise class representation \textbf{\textit{S}}:

\vspace{-4mm}
\begin{equation} 
w = S\times e, w \in R^{c\times 1}
\end{equation}

\begin{table*}[h]
\centering\small

\begin{tabular}{l||c|c|c|c}
 & \multicolumn{4}{c}{\bf Evaluation}  \\ \cline{2-5} & \bf Accuracy & \bf F1 & \bf R & \bf P  \\  \hline\hline
Naive Bayes    &     74.38    &     63.74    &     65.18  &     63.30   \\
RNN~\cite{lai2015recurrent}   &     77.10    &     64.40    &     63.90   &     64.84    \\ 
CNN~\cite{kim2014convolutional}  &     77.17    &     64.04   &     62.59   &     65.95   \\
EXAM~\cite{du2019explicit} &     77.34    &     62.75    &     60.80   & 66.36    \\
\hline
BERT-CLS & \underline{81.30}    &  68.82    & 67.26   &  71.11   \\
BERT-Avg-P  &     81.14    &     \underline{68.90}    &     \underline{67.50}   &     \underline{71.25} \\
\hline
CRAB-1 (Ours) &     81.51    &     68.85    &     66.98   &     71.80 \\
CRAB-2  (Ours) &     81.70    &     69.75    &     68.00   &     72.16 \\
CRAB-4 w/o SA  (Ours) &     81.86    &     69.97    &     68.31   &     72.27 \\
CRAB-4  (Ours) &    \bf  82.03 \small(RI: +0.9\%)   &   \bf   70.20  \small(RI: +1.89\%)  &    \bf  68.56  \small(RI: +1.57\%) &    \bf  72.54 \small(RI: +1.81\%)\\
\end{tabular}\

\caption{The performance of different models (in \%) on test set. F1: macro-averaged F1 score, P: macro-averaged Precision, R: macro-averaged Recall, and RI: relative improvement between bolded and underlined scores. All the improvements are statistically significant using a one-tailed Student’s t-test with a p-value $<$ 0.05.}
\label{results_eval}
\vspace{-6mm}
\end{table*}

\vspace{-4mm}

\subsection{Aggregation Layer}
\label{agg_lay}
This layer aims to fuse information learned from previous layers. Equations~\ref{agg1}-\ref{agg4} describe how we aggregate the token-level interaction signals and compute the matching scores, \textbf{\textit{z}}:

\vspace{-7mm}

\begin{eqnarray}
    T_i' &=& \sigma( T_i W_{fc1} + b_{fc1}) , i \in\{0,...,m\} \label{agg1}\\
    T''  &=& [T_0' \oplus ... \oplus T_m'] \label{agg2}\\
    V   &=& \sigma( T''W_{fc2} + b_{fc2})\label{agg3}\\
    z &=&  V W_{lin}, z\in R^{c\times 1} \label{agg4}
\end{eqnarray}

where $\sigma$ is LeakyReLu activation function~\cite{xu2015empirical} and $\oplus$ indicates concatenation operator. Also, multi-dimensional array $(T'_0,...,T'_m)$ is shown as \textbf{\textit{$T'$}} in Figure~\ref{fig:model}. In above equations, $W_{fc1},W_{fc2}$ and $W_{lin}$ are trainable weights. Finally, as denoted in equation~\ref{out}, token-level and sentence-level matching scores are combined and passed to a \textit{softmax} layer to predict the class labels:

\vspace{-3mm}

\begin{equation} 
\label{out}
o = \textit{softmax}(\frac{z}{|z|} + \frac{w}{|w|})
\end{equation}



\vspace{-2mm}
\section{Experiments}
\label{exp}
In this section, we describe our dataset, pre-processing steps, metrics, baseline models, hyper-parameter settings, and experimental procedures utilized to evaluate CRAB.
\subsection{Dataset}
We trained our model with tweets collected by \citeauthor{founta2018large}\footnote{https://sites.google.com/view/icwsm2020datachallenge} The corpus is comprised of 4 classes: \textit{Normal}, \textit{Abusive}, \textit{hateful} and \textit{Spam}. Class distribution is shown in Table~\ref{tbl:class_dist}; it is clear from the table that classes are heavily imbalanced. We apply stratified sampling and create train, validation and test sets with ratios of 80\%,10\%,10\% respectively.

\subsection{Data Pre-processing}
Tweets are full of emojis, emoticons, hashtags, and website links. To further clean tweets and yet preserve the useful information as much as possible, we develop a pipeline which maps emotionally similar emojis and emojicons into same special tokens. Likewise, the website links are also replaced with a special token. Generally, tweets are short and people do not follow grammatical rules in them. Therefore, we do not apply stop words removal, stemming or lemmatization as these techniques are often imperfect and could lead to information loss. 

\vspace{-2mm}
\subsection{Setup Details}
BERT was initialized with the pre-trained weights and we fine-tuned them for our downstream task during the training process. We chose batch size of 32 and number of neurons in the transformations $W_{fc1}$ and $W_{fc2}$ were 64 and 128 respectively. Model was implemented by Pytorch \cite{paszke2017automatic} with a single NVIDIA P100 GPU.

\vspace{-2mm}
\subsection{Baselines and Metrics}
As listed in Table \ref{results_eval}, we compare our approach with several baselines to evaluate effectiveness of our proposed model. The input to Naive Bayes classifier is TF-IDF \cite{zhang2011comparative} feature vectors. As part of our neural model baselines we employ CNN, RNN as well as EXAM \cite{du2019explicit}; the word embedding size of all aforementioned neural models are set to 200. We also include classification performances of feeding BERT's special token [CLS] embedding (BERT-CLS) and average pooled BERT's embedding vectors (BERT-Avg-P) to a linear classifier. In both cases the outputs of last hidden layer would be sent to the classifiers. Given that our class distribution is imbalanced, we consider Macro-Averaged F1, Precision, Recall as well as Accuracy to quantify the prediction performance of the models.

\vspace{-2mm}
\subsection{Empirical Results and Discussion}

Table~\ref{results_eval} shows performances of different variations of our model, CRAB-n, as well as the baselines. Bold numbers denote the best performance and underlined numbers are the best performance among the baselines only. Letter \textbf{\textit{n}} in CRAB-n indicates number of heads in token-wise class representation layer. During our experiment we tried various $n\in\{1,2,4,8,16\}$. To evaluate effectiveness of sentence-wise class representation layer, we also report performance of CRAB-4 with this layer removed (depicted as CRAB-4 w/o SA in Table~\ref{results_eval}). As shown in Table~\ref{results_eval}, CRAB-4 consistently outperformed all baselines and other CRAB variations. Our model, CRAB-4, achieved \textbf{1.89\%} relative improved Macro-averaged F1 score and \textbf{0.9\%} relative improved accuracy compared to BERT-Avg-P and BERT-CLS respectively. In terms of Macro-averaged precision and recall, there is relative improvement of \textbf{1.81\%} and \textbf{1.57\%} over BERT-Avg-P, respectively. To further analyze performance of CRAB-4 and understand the way it handles imbalanced classes, we conducted error analysis on each class. Compared to our baseline BERT models, we noticed CRAB-4 obtained 1\% Macro-F1 boost in the first two major classes and for the two minor classes it gained 2\% increase. We hypothesize that,  minor classes benefited the most from this architecture. It is worth mentioning our model can be extended to multi-label classification by simply replacing \textit{softmax} with \textit{sigmoid} layer. We emphasize that in our proposed architecture, input representation layer is not just limited to BERT and any other form of transformer-based encoder can be used instead. 

\vspace{-1mm}
\section{Conclusion and Future Work}
\vspace{-2mm}

In this paper, we introduced CRAB, a neural model to identify hate speech from Twitter data. CRAB incorporates both word and class information from tweets into the hate speech identification process. CRAB significantly outperformed the state-of-the-art BERT-based baseline by 1.89\% on relative Macro F1. Our future work includes evaluating effectiveness of this model in the extreme multi-class and multi-label problems and adopting CRAB for other online abusive behavior detection tasks.


\bibliography{anthology,emnlp2020}
\bibliographystyle{acl_natbib}

\end{document}